\documentclass[10pt, conference, letterpaper]{IEEEtran}
\IEEEoverridecommandlockouts
\usepackage{cite}
\usepackage{amsmath,amssymb,amsfonts}
\usepackage{algorithmic}
\usepackage{graphicx}
\usepackage{textcomp}
\usepackage{xcolor}
\usepackage{booktabs} 
\usepackage{algorithm} 
\usepackage{multirow}
\usepackage{hyperref}

\usepackage{capt-of}
\usepackage{array}
\usepackage{makecell}

\newcolumntype{L}[1]{%
  >{\raggedright\arraybackslash}p{#1}}

\usepackage[normalem]{ulem}
\useunder{\uline}{\ul}{}

\def\BibTeX{{\rm B\kern-.05em{\sc i\kern-.025em b}\kern-.08em
    T\kern-.1667em\lower.7ex\hbox{E}\kern-.125emX}}
\begin{document}

\title{CSI-Agent: LLM-Assisted Few-Shot Adaptation for Cross-Domain Wi-Fi CSI Sensing}

\author{\IEEEauthorblockN{Tianya Zhao$^*$, Chuan Liu, Xuyu Wang\textsuperscript{* \textsection}}
\IEEEauthorblockA{
$^*$Knight Foundation School of Computing and Information Sciences, Florida International University, Miami, FL 33199, US\\
Emails: tzhao010@fiu.edu, cliu060@fiu.edu, xuywang@fiu.edu}
}

\maketitle


\begin{abstract}
Wi-Fi channel state information (CSI) has enabled device-free sensing applications such as human activity recognition. However, CSI sensing models remain brittle in cross-domain deployment, where changes in users or environments can produce incorrect predictions. Existing solutions usually treat this problem as an offline model-design problem, by pretraining a stronger representation or applying one fixed adaptation method to the entire target domain. In practice, labeled target data are scarce and different classes may fail in different ways under the same domain shift.
To address this, we propose CSI-Agent, an evidence-seeking LLM agent that reformulates cross-domain CSI adaptation as a deployment-time decision-making problem. Rather than processing raw CSI or making sample-level predictions, CSI-Agent summarizes target-domain behavior into sensing-grounded class-level evidence. It establishes a strong target-adaptive default from complementary CSI views and uses an LLM planner to determine whether each class should retain the default or invoke a specialized action. Deterministic verification and bounded execution further reduce unreliable interventions.
We evaluate CSI-Agent on four public datasets using five cross-domain splits covering device, user, environment, and compositional shifts. Under 1-shot adaptation, CSI-Agent achieves the best target-domain performance across all splits and improves the average Macro-F1 by about 16\% compared to the strongest baseline method.
\end{abstract}


\section{Introduction}\label{sec:intro}

Wi-Fi channel state information (CSI) has emerged as an important sensing modality for ubiquitous and mobile computing~\cite{li2026uni, zhao2024functional, wang2017phasebeat, he2023sencom}. Wi-Fi sensing is particularly attractive because it requires no dedicated wearable devices, avoids capturing visual content, and can leverage existing commodity wireless infrastructure. By characterizing how wireless signals propagate through an environment, CSI captures multipath variations induced by human motion, body posture, gestures, and interactions with surrounding objects~\cite{ma2019wifi}. Recent advances in deep learning have substantially improved the performance of CSI-based applications, including human activity recognition~\cite{jiang2018towards}, fall detection~\cite{wang2016rt}, and gesture recognition~\cite{ma2018signfi}.

Despite this progress, CSI-based sensing remains fragile under cross-domain deployment~\cite{chen2023cross}. In this work, a domain is characterized by deployment-specific factors that affect the CSI distribution, such as the environment, user, and transceiver configuration. A domain shift occurs when these factors differ between training and deployment, even though the sensing task remains unchanged. For example, a CSI model trained in one room may suffer substantial performance degradation when deployed in another. This fragility arises because CSI is jointly shaped by human motion, environmental geometry, transceiver configuration, hardware characteristics, and multipath propagation. Consequently, the same activity may produce substantially different CSI patterns across domains, while different activities may become less distinguishable in an unseen domain.

Existing approaches address domain shift at different stages of the sensing pipeline. Before deployment, supervised training across multiple domains and self-supervised pretraining aim to learn more transferable representations~\cite{chen2020simple,hong2024crosshar}. During deployment, domain adaptation leverages target-domain observations to align source and target distributions~\cite{farahani2021brief,ganin2016domain,jiang2018towards}, while few-shot adaptation uses a small number of labeled target samples to specialize the model to a new domain~\cite{zhao2024cross,yin2022fewsense,snell2017prototypical}. These approaches have substantially improved cross-domain generalization, but they make different assumptions about the target-domain data available at deployment.

Collecting a large labeled CSI dataset for every new deployment is costly and often impractical. Few-shot target-domain adaptation therefore offers a practical compromise: limited labeled examples may be collected, for example, through a brief user-enrollment or site-calibration procedure~\cite{wang2026survey,feng2022wi, wang2024review}. These examples provide valuable evidence about the new domain without requiring full-scale data collection and massive model retraining. After this initial setup, the sensing system should operate autonomously on subsequent unlabeled observations without repeatedly requesting additional annotations.

However, existing few-shot adaptation methods typically follow a predetermined adaptation recipe and apply it uniformly across the target domain. Such a fixed treatment overlooks the heterogeneous effects of domain shift and adaptation across classes. A mechanism may improve a large portion of classes, have little effect on others, and further degrade those for which it is poorly matched. 
Consequently, no single mechanism is uniformly beneficial across the target domain, and its global application may introduce negative transfer. Selecting different actions for different data is also non-trivial, because cross-domain models can remain highly confident even when their predictions are incorrect.

These limitations motivate a different view of cross-domain Wi-Fi CSI sensing systems. Rather than treating adaptation solely as the optimization of a fixed model or the execution of a predetermined action, we formulate it as a deployment-time decision-making problem. Under this formulation, the sensing system must diagnose specific domain shift behavior, determine whether and how recovery should be performed, and reject actions that may degrade sensing performance. 
Therefore, the focus shifts from learning a single adapted model to controlling a set of complementary sensing actions under limited target-domain evidence.

This deployment-time decision process naturally calls for an agentic controller. Large language model (LLM) agents provide a promising foundation for such a controller because they can integrate heterogeneous structured evidence, coordinate specialized tools, and provide explicit rationales for their decisions~\cite{luo2025large,yao2022react,schick2023toolformer,huang2025modality}. Rather than applying one single adaptation rule uniformly, an LLM agent can reason about the observed deployment conditions and construct a class-conditioned recovery plan. Moreover, the same high-level planner can be applied across users, environments, and transceiver configurations without retraining for each new domain shift.

\textbf{Challenges.} 
Realizing this LLM-agent-based approach for CSI sensing presents three key challenges.
First, raw CSI measurements are high-dimensional, non-linguistic signals that are unsuitable for reliable interpretation by general-purpose LLMs. The agent must reason from sensing-grounded evidence that preserves the numerical and physical characteristics relevant to domain shift.
Second, the available target-domain evidence is limited and uncertain. Labeled data from the new domain are scarce and costly to obtain, while most target-domain samples remain unlabeled. Different diagnostic signals may also provide conflicting indications about the behavior of individual classes. Therefore, the agent must determine whether adaptation is necessary and which recovery action is appropriate under this uncertainty.
Third, an adaptation action that appears promising based on limited evidence may still degrade certain classes or reduce overall sensing performance. Therefore, each proposed action must be evaluated before execution, while a reliable fallback should be retained when the evidence is insufficient.

\textbf{Solution.} To address these challenges, we develop CSI-Agent, an LLM-agent-based system for cross-domain CSI adaptation at deployment time.
First, specialized sensing tools process raw CSI and convert it into structured numerical and physical evidence, including the reliability of limited labeled target data, class-wise activation and false-activation statistics, and the relative behavior of complementary signal views.
Second, the LLM agent integrates this evidence with statistics derived from unlabeled target observations to diagnose transfer behavior for each class and construct a class-conditioned adaptation plan.
Third, a sensing-grounded verifier evaluates every proposed action using reliability and no-harm criteria and replaces unreliable actions with a safer fallback. In this design, the LLM does not directly classify CSI. Instead, it provides the control intelligence needed to reason under limited target-domain supervision, coordinate deterministic sensing tools, and produce auditable adaptation decisions.

The main contributions of this paper are as follows.
\begin{itemize}
    \item To the best of our knowledge, this is the first work to investigate LLM-agent-based deployment-time adaptation for cross-domain Wi-Fi sensing. We reformulate adaptation from a fixed model-optimization procedure into a decision-making problem with practical constraints.
    \item We propose CSI-Agent, a sensing-grounded agentic framework that separates high-level reasoning from CSI signal processing. Specialized tools extract structured diagnostic evidence, an LLM agent constructs a class-conditioned adaptation plan, and a verifier evaluates the proposed actions using reliability and no-harm criteria before execution.
    \item We evaluate CSI-Agent on four public Wi-Fi CSI datasets using five cross-domain splits covering device, user, environment, and compositional shifts. The results demonstrate that CSI-Agent improves cross-domain sensing performance over fixed adaptation baselines, reduces negative transfer, and produces auditable adaptation decisions.
\end{itemize}

The rest of the paper is organized as follows. Section~\ref{sec:bg} provides background and motivation. Section~\ref{sec:relatedwork} discusses related work. Section~\ref{sec:formulation} illustrates the problem formulation. Section~\ref{sec:method} presents our CSI-Agent design. Section~\ref{sec:ex_eva} evaluates the proposed system. Section~\ref{sec:conclusion} concludes the paper.

\section{Background and Motivation}\label{sec:bg}

\subsection{Channel State Information}

In OFDM-based Wi-Fi systems, CSI estimates the complex channel response between a transmitter and receiver over multiple subcarriers and antenna links. For subcarrier $m$ at time $t$, a CSI coefficient can be written as $H_t(m)=A_t(m)e^{j\phi_t(m)}$, where $A_t(m)$ and $\phi_t(m)$ denote the measured amplitude and phase, respectively. The received signal contains multiple propagation paths caused by reflection, diffraction, and scattering.
Human motion changes the lengths and strengths of these paths, producing temporal variations in CSI that encode activity-related information.
By collecting CSI over consecutive packets, a sensing system obtains a time-frequency representation that can be used for activity recognition~\cite{ma2019wifi}.

\subsection{Preliminary Study}\label{sec:pre}
We conduct preliminary experiments on CSI-Bench~\cite{zhu2026csi} to further motivate deployment-time adaptation for cross-domain Wi-Fi CSI sensing. 
The study considers a practical few-shot deployment setting in which a source-trained sensing model and a small labeled target set are available, while the remaining target samples are unlabeled. 
We first examine whether a fixed adaptation strategy is uniformly reliable across target domains and classes.
We then investigate whether reliable LLM planning requires structured, sensing-grounded evidence rather than source-model confidence or raw CSI.

\begin{figure}[th]
    \centering
    \includegraphics[width=1\columnwidth]{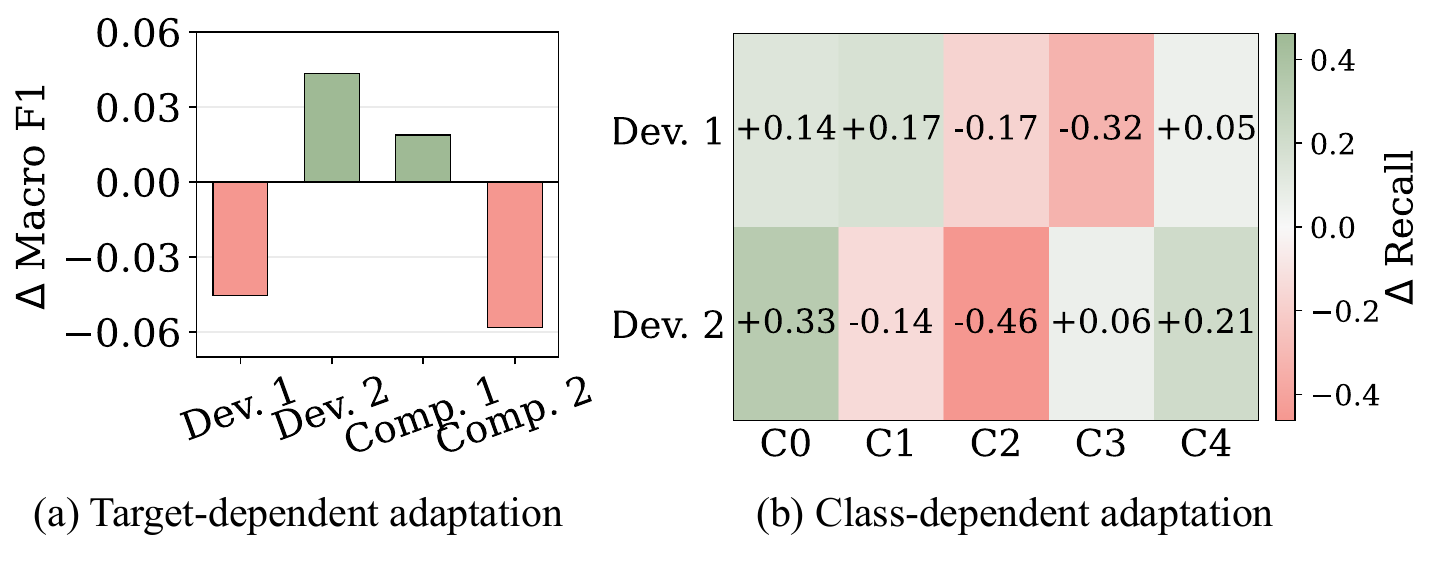}
    \caption{Fixed few-shot adaptation is not uniformly reliable. Dev.~1 and Dev.~2 denote EchoSpot and AmazonPlug, while Comp.~1 and Comp.~2 denote U02/E01 and U05/E05 in CSI-Bench~\cite{zhu2026csi}, respectively.
    (a) A fixed 1-shot PTN can either improve or degrade different target domains. (b) Within the same cross-device split, PTN affects activity classes unevenly.}
    \label{fig:motivation_1}
\end{figure}

\textbf{Observation 1: Fixed adaptation is target-dependent.}
We use a standard 1-shot prototypical network (PTN)~\cite{snell2017prototypical} as a representative fixed few-shot adaptation strategy and compare it with the source ResNet~\cite{he2016deep}.
A natural use of limited target labels is to apply the same adaptation strategy to every target domain. However, Fig.~\ref{fig:motivation_1}(a) shows that this fixed adaptation is not uniformly reliable. The 1-shot PTN improves macro-F1 on Dev.~2 and Comp.~1, while reducing it on others.
These results show that limited target labels can be valuable, but a predetermined adaptation strategy may also introduce negative transfer when applied uniformly.

\textbf{Observation 2: Adaptation effects are class-dependent.}
The effect of adaptation also varies across sensing classes within the same target domain. Fig.~\ref{fig:motivation_1}(b) reports the class-wise recall changes from the source model to the 1-shot ProtoNet on two cross-device target domains.
Under the same adaptation strategy, some classes improve substantially, while others experience considerable degradation. A single domain-level decision is therefore too coarse to capture heterogeneous class-level transfer behavior. 
Deployment-time adaptation should instead diagnose individual classes and allow them to use different adaptation actions.

\begin{figure}[th]
    \centering
    \includegraphics[width=1\columnwidth]{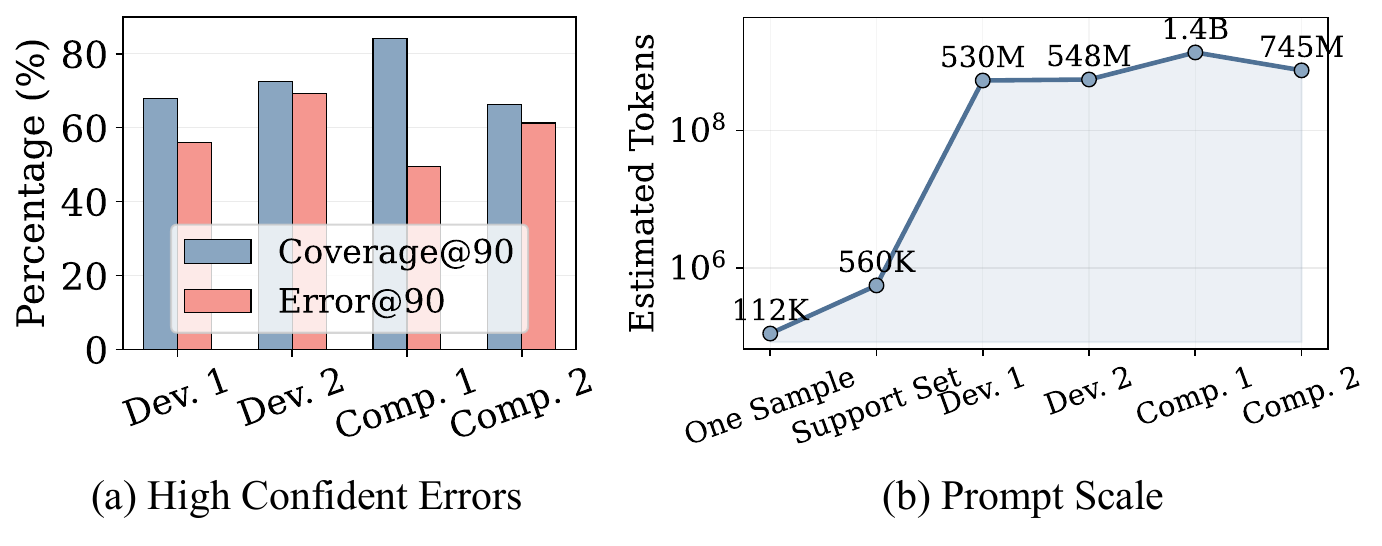}
    \caption{Raw confidence and raw CSI are unsuitable for deployment-time LLM planning.
    (a) On target domains, source-model predictions with confidence above 0.9 still contain many errors, showing that confidence alone is an unreliable control signal.
    (b) Directly serializing CSI samples into an LLM prompt leads to a rapidly growing token scale, motivating compact sensing-grounded evidence before LLM planning.}
    \label{fig:motivation_2}
\end{figure}

\textbf{Observation 3. LLM planning requires structured sensing evidence.}
A natural question is whether readily available deployment signals can be used directly for LLM planning. We examine two intuitive choices: source-model confidence and raw CSI data.
As shown in Fig.~\ref{fig:motivation_2}(a), among source predictions with confidence above 0.9, 49.6\% to 69.2\% are still incorrect. Confidence alone is therefore insufficient for diagnosing transfer failures or selecting adaptation actions.
Meanwhile, Fig.~\ref{fig:motivation_2}(b) shows that serializing one CSI sample already requires approximately 112K tokens, while a target-domain query set requires 530M to 1.4B tokens. Raw CSI is thus impractical as direct LLM input and does not explicitly expose the class-level behavior needed for adaptation.

\textbf{Motivation.}
These observations lead to three design requirements for CSI-Agent.
First, adaptation should account for target-specific effects, since the same few-shot strategy may improve one target domain while degrading another. Importantly, this does not imply that a strong domain-level adaptation method is broadly ineffective. It may perform well overall while remaining suboptimal for individual classes.
Second, adaptation should operate at the class level, as different sensing classes under the same domain shift may require different treatments.
Third, LLM planning should be grounded in structured sensing evidence rather than relying solely on readily available deployment inputs, such as raw CSI and source-model confidence.

\section{Related Work}\label{sec:relatedwork}

\subsection{Cross-Domain Wi-Fi Sensing}
Cross-domain Wi-Fi sensing systems aim to maintain sensing performance under changes in users, environments, and transceiver configurations~\cite{chen2023cross}. Early studies primarily improve generalization by learning domain-invariant representations or constructing more robust physical features.
CrossSense trains a roaming model to generate synthetic samples for new environments, thereby reducing target-site data collection costs~\cite{zhang2018crosssense}.
EI employs adversarial learning to remove environment- and user-specific information from activity representations~\cite{jiang2018towards}, while Widar 3.0 derives a body-coordinate velocity profile from multiple Wi-Fi links for gesture recognition~\cite{zhang2021widar3}.
WiHF derives a domain-independent motion-change pattern for real-time gesture recognition and user identification~\cite{li2020wihf}.

Recently, more studies reduce target-domain labeling costs through one-shot or few-shot adaptation.
TOSS performs semi-supervised domain adaptation by jointly leveraging a small number of labeled target samples and abundant unlabeled target observations~\cite{zhou2022target}.
Wi-Learner extracts Doppler features and adapts to a new deployment using one labeled example per gesture~\cite{feng2022wi}. 
FewSense adapts a pretrained feature extractor using a few target-domain samples and performs similarity-based recognition~\cite{yin2022fewsense}.
MetaFormer further combines meta-learning with Transformer representations to adapt using only one labeled target sample per class~\cite{sheng2024metaformer}.
OneSense employs an aug-meta learning framework to enable scalable few-shot learning for gesture recognition systems~\cite{zhao2024one}.
Although these methods require target-domain supervision, they generally employ a predetermined adaptation procedure after deployment.

\subsection{LLM-Based Sensing}
Recent studies have explored large language models for wireless sensing and signal processing.
Wi-Chat incorporates physical knowledge of Wi-Fi sensing into prompts and uses an LLM to perform zero-shot activity recognition~\cite{zhang2025wi}.
AutoIoT translates natural-language user requirements into interpretable programs for AIoT applications and iteratively improves the generated programs with limited user intervention~\cite{shen2025autoiot}.
SignalLLM adopts an agentic approach to decompose general signal-processing tasks, construct processing workflows, and invoke external tools~\cite{ke2025signalllm}.


Our work advances these directions by reframing cross-domain adaptation from a predetermined procedure into an evidence-driven and verifiable decision process. CSI-Agent grounds an LLM planner in structured target-domain diagnostics to determine whether each class should retain a strong default or use a specialized sensing action, while deterministic tools handle CSI processing, prediction, verification, and bounded execution.


\section{Problem Formulation}\label{sec:formulation}

In this paper, we consider closed-set Wi-Fi CSI sensing under cross-domain deployment. Let $\mathcal{Y}=\{1,\ldots,C\}$ denote the known label space for a sensing task. A labeled source dataset $\mathcal{D}^{src}$ is available before deployment and is used to train a source sensing model $f_{\theta}$. 
At deployment time, $f_{\theta}$ is applied to a target domain $d^{tar}$ whose CSI distribution may differ from the source because of new users, environments, or devices, while the label space $\mathcal{Y}$ stays the same.

For each target domain, the system observes a small labeled support set as follows:
\begin{equation}
    \mathcal{S}^{tar} = \{(x_i^{s},y_i^{s})\}_{i=1}^{Ck},
\end{equation}
with $k$ samples per class, which may be collected through a brief enrollment or calibration procedure.
The system also observes an unlabeled target set:
\begin{equation}
    \mathcal{Q}^{tar} = \{x_j^{q}\}_{j=1}^{N_q},
\end{equation}
whose ground-truth labels need to be inferred during deployment. 
Unlike the small episodic query sets used in conventional few-shot learning, $\mathcal{Q}^{tar}$ covers every unlabeled sample in the target domain, so $N_q$ is typically much larger than $Ck$.

Instead of learning a single adapted model, we formulate deployment-time adaptation as a decision problem.
The system selects a target-domain predictor $h$ from a constrained family $\mathcal{H}$ using only the information above. The ideal predictor maximizes target-domain macro-F1:
\begin{equation}
    h^{*}=\arg\max_{h\in\mathcal{H}}\;
    \mathrm{MacroF1}\big(\{y_j^{q}\},\{h(x_j^{q})\}\big),
\end{equation}
where $\{y_j^q\}$ denotes the hidden query labels used only for offline evaluation. During deployment, CSI-Agent selects $h$ using only labeled support behavior and unlabeled query statistics.
In this paper, CSI-Agent instantiates $\mathcal{H}$ through class-conditioned combinations of deterministic sensing actions.

\section{System Design}\label{sec:method}

\subsection{Overview}

\begin{figure}[th]
    \centering
    \includegraphics[width=1\columnwidth]{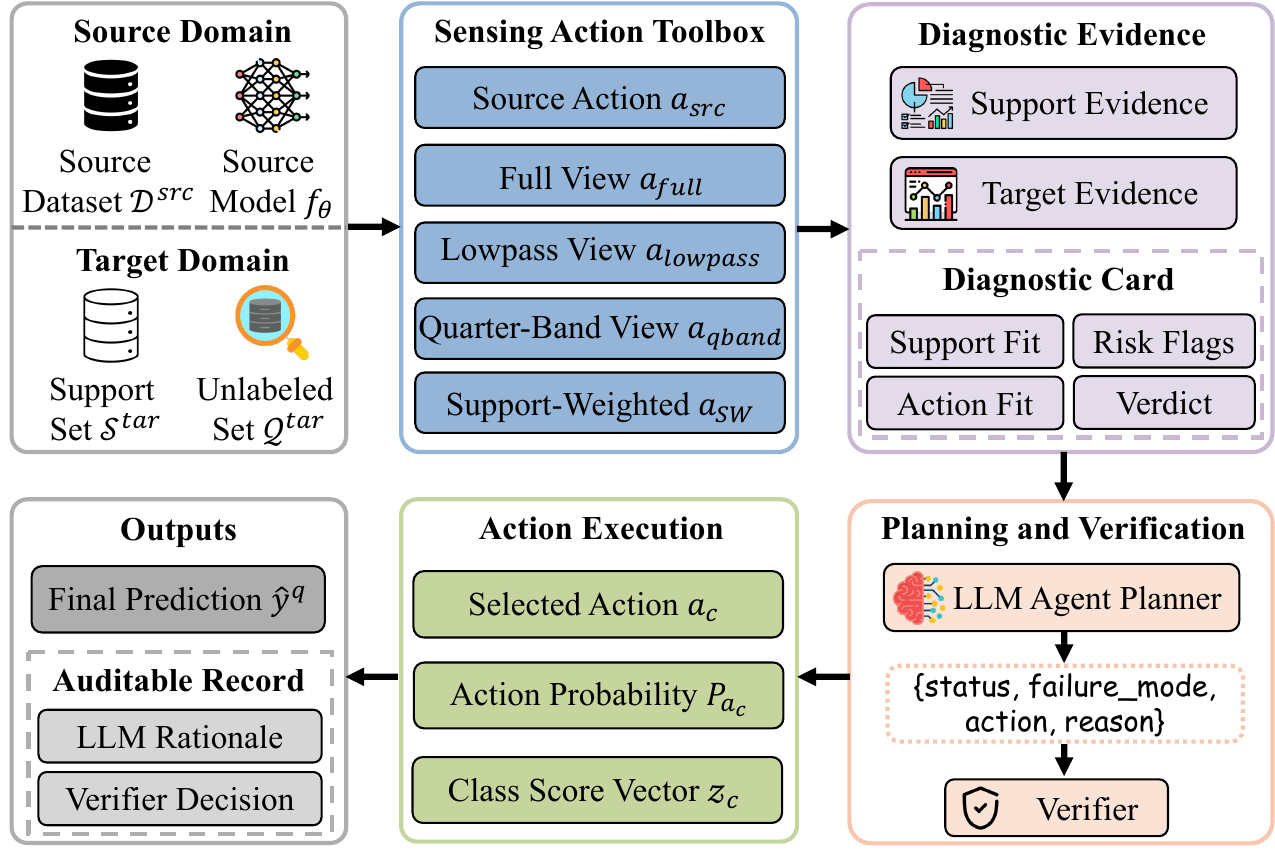}
    \caption{Overview of the CSI-Agent framework.}
    \label{fig:overview}
\end{figure}

Fig.~\ref{fig:overview} presents the overview of CSI-Agent. 
Given a source sensing model, a small labeled target support set, and an unlabeled target query set, CSI-Agent performs deployment-time adaptation through four stages. 
It first constructs a constrained toolbox of complementary deterministic sensing actions. It then summarizes their class-level behavior into sensing-grounded diagnostic cards. Based on this evidence, an LLM agent proposes an action for each class, while a deterministic verifier rejects unreliable proposals. Finally, a bounded executor applies the accepted actions relative to a common support-weighted default and produces the target-domain predictions.

\subsection{Sensing Action Toolbox}


CSI-Agent uses a constrained set of deterministic sensing actions as follows:
\begin{equation}
    \mathcal{A}
    =
    \{
    a_{\mathrm{src}},
    a_{\mathrm{full}},
    a_{\mathrm{lowpass}},
    a_{\mathrm{qband}},
    a_{\mathrm{SW}}
    \}.
\end{equation}
We use the term action to denote a fixed sensing scorer rather than a free-form program. Each action follows a predefined execution rule and produces a class-probability vector.

\textbf{Source action.} The source action directly applies the frozen source classifier $f_{\theta}$ to the original CSI sample:
\begin{equation}
    P_{\mathrm{src}}(x)=\operatorname{softmax}(f_{\theta}(x)).
\end{equation}
This action remains effective when the source representation transfers well to the target domain, but it may degrade substantially under severe domain shifts.

\textbf{View-specific recovery actions.} To capture different features, CSI-Agent uses three amplitude-based CSI views $\mathcal{V} \in \{\mathrm{full}, \mathrm{lowpass}, \mathrm{qband}\},$ where $v\in\mathcal{V}$ indexes a CSI view.
The $\mathrm{full}$ view retains the original CSI representation. The $\mathrm{lowpass}$ view applies a short temporal moving average to suppress unstable high-frequency variations. 
The quarter-band view, denoted by $\mathrm{qband}$, retains a contiguous quarter of the subcarriers and masks the remaining ones, thereby providing a localized spectral view.

Let $T_v$ denote the input transformation associated with view $v$, and let $g_{\theta}$ denote the feature extractor of the source-trained model. The view-specific representation is
\begin{equation}
    h_v(x)=g_{\theta}(T_v(x)).
\end{equation}
For each view $v\in\mathcal{V}$ and class $c$, CSI-Agent computes source prototypes as follows:
\begin{equation}
    \mu_c^v = \frac{1}{|\mathcal{D}^{src}_c|}
    \sum_{(x_i^{\rm src},y_i^{\rm src})\in\mathcal D_c^{\rm src}} h_v(x_i^{\rm src}),
\end{equation}
where $\mathcal{D}^{src}_c$ contains the source samples from class $c$.
Then, CSI-Agent estimates a coarse target-domain feature shift without query labels:
\begin{equation}
    \Delta^v = \frac{1}{|\mathcal{T}|}\sum_{x\in \mathcal{T}} h_v(x)
    - \frac{1}{|\mathcal{D}^{src}|}\sum_{x\in \mathcal{D}^{src}} h_v(x),
\end{equation}
where $\mathcal{T}=\mathcal{S}\cup\mathcal{Q}$ denotes the available target-domain batch. 
The shifted source prototype is $\widetilde{\mu}_c^v = \mu_c^v + \Delta^v.$
This shared shift provides coarse alignment for global differences between the source and target feature distributions. However, a single global shift may not align all classes equally well, while estimating reliable class-specific shifts from only a few labeled samples is difficult. CSI-Agent therefore evaluates the resulting scorers at the class level and uses this evidence for subsequent action selection.

CSI-Agent uses the shifted prototypes $\widetilde{\mu}_c^v$ in two complementary ways. First, it converts the distances between labeled support samples and shifted prototypes into support-side diagnostic probabilities:
\begin{equation}
    P_v^S(x_i^{s},c)=
    \frac{
        \exp\left(-\|h_v(x_i^{s})-\widetilde{\mu}_c^v\|_2^2/\tau\right)
    }{
        \sum_{c'\in\mathcal{Y}}
        \exp\left(-\|h_v(x_i^{s})-\widetilde{\mu}_{c'}^v\|_2^2/\tau\right)
    },
\end{equation}
where $\tau>0$ is a temperature parameter, set to $32$ in all experiments. 
A higher $P_v^S(x_i^{s},c)$ indicates that the view-specific representation $h_v(x_i^{s})$ is closer to the shifted prototype of class $c$ relative to the competing class prototypes. These probabilities are used only to evaluate the reliability of view $v$ on the labeled support set.

Second, to obtain query predictions while preserving source-domain class structure, CSI-Agent fits a ridge classifier. Let $\phi(u)=[u^\top,1]^\top$ denote the bias-augmented version of feature vector $u$. For each view $v$, the classifier is obtained by
\begin{equation}
\begin{split}
    W_v^* = \arg\min_W
    &\sum_{(x_i^s,y_i^s)\in\mathcal{S}}
    \left\|
        \phi(h_v(x_i^s))^\top W-\mathbf{e}_{y_i^s}^\top
    \right\|_2^2 \\
    &+\lambda\sum_{c=1}^{C}
    \left\|
        \phi(\widetilde{\mu}_c^v)^\top W-\mathbf{e}_c^\top
    \right\|_2^2
    +\alpha\|W\|_F^2,
\end{split}
\end{equation}
where $W$ maps the augmented feature representation to $C$ class scores, $\mathbf{e}_c\in\mathbb{R}^{C}$ is the one-hot vector of class $c$, $\lambda\geq0$ controls the prototype anchoring strength, and $\alpha>0$ controls the ridge regularization, with $\|W\|_F^2=\sum_{i,j}W_{ij}^2$. In this paper, we set $\lambda=1$ and $\alpha=1$.
The first term fits the classifier to the limited labeled target samples, while the second term retains the class structure represented by the shifted source prototypes. 
For an unlabeled query sample $x_j^q\in\mathcal{Q}$, the view-specific probability vector is
\begin{equation}
    P_v^Q(x_j^q)
    = \operatorname{softmax} \left(\phi(h_v(x_j^q))^\top W_v^*\right).
\end{equation}

\textbf{Support-weighted action.} To reduce reliance on any single CSI view, CSI-Agent combines the three view-specific scorers according to their reliability on the labeled support set. For each view $v$, let $\widehat{y}_{v,i}^{s} = \arg\max_{c\in\mathcal{Y}} P_v^S(x_i^s,c)$ denote the predicted label of support sample $x_i^s$. The support reliability of view $v$ is defined as
\begin{equation}
    r_v = \max
    \left(
        \operatorname{MacroF1}
        \left(
            \{y_i^s\}_{i=1}^{Ck},
            \{\widehat{y}_{v,i}^{s}\}_{i=1}^{Ck}
        \right),
        \epsilon
    \right),
\end{equation}
where $\epsilon>0$ is a small constant that prevents a view from receiving
an exactly zero weight. The normalized weight of view $v$ is $w_v = \frac{r_v} {\sum_{u\in\mathcal{V}}r_u}.$
For an unlabeled query sample $x_j^q$, the support-weighted probability vector is
\begin{equation}
    P_{\mathrm{SW}}^Q(x_j^q) =
    \sum_{v\in\mathcal{V}}w_vP_v^Q(x_j^q).
\end{equation}
The support-weighted action provides a target-adaptive and robust default. It emphasizes views that are more reliable on the labeled target samples while retaining complementary evidence from the other views. 

\subsection{Sensing-Grounded Diagnostic Evidence}

For each known class $c\in\mathcal{Y}$, CSI-Agent collects diagnostic
evidence that summarizes how each action behaves on the labeled support set and
the unlabeled query set.

\textbf{Support evidence.}
For the positive support samples with $y_i^s=c$, CSI-Agent measures the class-wise recall, the mean probability assigned to class $c$, and the mean rank of class $c$. These quantities characterize how reliably an action recognizes the class. For the negative support samples with $y_i^s\neq c$, CSI-Agent measures the false-activation rate and computes a one-vs-rest margin between the average class scores on positive and negative samples. 
Together, these measurements capture both recognition of class $c$ and interference from competing classes.

\textbf{Unlabeled target evidence.}
Since the query labels are unavailable, CSI-Agent summarizes only the prediction behavior induced by each action. For action $a\in\mathcal{A}$ and class $c\in\mathcal{Y}$, the primary statistic is the query activation rate:
\begin{equation}
    \mathrm{QRate}_{a,c} = \frac{1}{N_q}
    \sum_{j=1}^{N_q} \mathbb{I}
    \left[
        \arg\max_{c'\in\mathcal{Y}} P_a^Q(x_j^q,c') = c
    \right].
\end{equation}
This quantity represents the fraction of unlabeled target samples assigned to class $c$ by action $a$. 
CSI-Agent compares $\mathrm{QRate}_{a,c}$ with the activation rate of the support-weighted default, denoted by $\mathrm{QRate}_{\mathrm{SW},c}$, to determine how the candidate action changes the target-domain prediction pattern.
In this paper, we use $\rho_c=\frac{1}{C}$ as a coarse uniform reference. This reference makes the weak assumption that the target classes occur at roughly similar frequencies. It is used only to flag severe under-activation or over-activation, rather than to estimate the true target-domain class distribution, and is never used alone to accept an action.

\textbf{Diagnostic card.}
CSI-Agent organizes the support evidence and query activation statistics into
a structured diagnostic card for each class and candidate action. 
The card contains four fields:
\begin{itemize}
    \item \emph{Support Fit}, which indicates whether the action improves,
    preserves, or worsens class separation on the labeled support samples.
    \item \emph{Activation Fit}, which indicates how the action changes the
    target-domain activation of the class relative to the default scorer and
    the coarse uniform reference.
    \item \emph{Risk Flags}, which identify increased false activation,
    aggravated under-activation or over-activation, and conflicts between
    support and unlabeled target evidence.
    \item \emph{Verdict}, which summarizes the action as viable, weak, or risky
    before LLM planning.
\end{itemize}

\subsection{LLM Agent Planning and Verification}

Given the diagnostic cards constructed in the previous stage, CSI-Agent uses an LLM to propose a sensing action $a_c\in\mathcal{A}$ for each class $c$. A deterministic verifier then evaluates each proposed action before execution.

\textbf{LLM agent planning.}
For each class $c\in\mathcal{Y}$, the LLM agent receives its diagnostic card together with a set of candidate actions. Although a general-purpose LLM may encode knowledge about wireless sensing, it is not directly grounded in high-dimensional, non-linguistic CSI measurements. CSI-Agent therefore does not expose raw CSI to the LLM or ask it to produce sample-level predictions. 
Instead, the agent diagnoses the observed transfer behavior and proposes a deterministic scorer based on the sensing-grounded evidence.
The planner returns a structured output as follows:
\begin{equation}
    o_c =
    \{\mathrm{status}, \mathrm{failure\_mode}, \mathrm{action}, \mathrm{reason}\}.
\end{equation}
The status indicates whether class $c$ appears reliably transferred or requires intervention. The failure mode summarizes whether the class is under-activated, over-activated, confused with competing classes, or poorly separated. The action field records the proposed action $a_c$, while the reason provides an auditable explanation grounded in the diagnostic card.

\textbf{Sensing-grounded verification.}
The LLM agent provides the primary diagnosis and action proposal. Since the available target-domain evidence is limited, CSI-Agent applies a lightweight deterministic verifier before execution. The verifier checks whether the proposed action is consistent with the diagnosed failure mode and whether it provides sufficient evidence of improvement over the support-weighted default without introducing substantial false activation or degrading support-side class separation.

If these conditions are not satisfied, the proposal is rejected and the class retains the support-weighted action. The verifier decision is represented by $g_c\in\{0,1\}$, where $g_c=1$ accepts the proposed action $a_c$ and $g_c=0$ activates the default fallback. Candidate filtering limits clearly implausible options before planning, whereas verification evaluates the final LLM proposal before execution. The action $a_c$ and gate $g_c$ are then passed to the bounded executor.

\subsection{Bounded Action Execution}

Following verification, CSI-Agent applies each accepted action as a bounded adjustment to the common support-weighted default. For query sample $x_j^q$ and class $c$, the final class score is
\begin{equation}
    z_c(x_j^q) = P_{\mathrm{SW}}^Q(x_j^q,c)
    + \beta g_c
    \left[P_{a_c}^Q(x_j^q,c) - P_{\mathrm{SW}}^Q(x_j^q,c)
    \right],
\end{equation}
where $\beta\in[0,1]$ controls the intervention strength and is set to $0.35$. A rejected action retains the support-weighted score, while an accepted action modifies it without necessarily replacing the default completely.
Then, the final prediction is 
$\widehat{y}_j^q=\arg\max_{c\in\mathcal{Y}} z_c(x_j^q).$
This bounded execution preserves a common scoring reference across classes while limiting negative transfer from uncertain action proposals.

\section{Experimental Evaluation}\label{sec:ex_eva}

\subsection{Experiment Setup}

All experiments are conducted on a Linux server with an Intel(R) Xeon(R) Gold 6258R CPU and NVIDIA A100 GPUs with 40GB of memory.
All source models are optimized with AdamW at a learning rate of \(3\times10^{-4}\) for at most 200 epochs, with early stopping (patience 10) after five epochs. SimCLR uses 200 pretraining and linear-probe epochs. EI and TOSS use learning rates of \(10^{-4}\) and \(5\times10^{-5}\), respectively.
For the main CSI-Agent experiments, the LLM is Qwen3.5-9B.

\textbf{Datasets and splits.}
We evaluate CSI-Agent on four public Wi-Fi CSI datasets using five cross-domain splits that cover device, user, environment, and compositional shifts, as summarized in Table~\ref{tab:datasets}.
For CSI-Bench~\cite{zhu2026csi}, we follow the official Cross-Device split and the Compositional split containing the held-out user--environment pairs U02/E01 and U05/E05. 
For WiAR~\cite{guo2019wiar}, users a--d form the source domain, while users e, f, 7, 8, 9, and 10 define six target domains.
For SignFi~\cite{ma2018signfi}, we retain the first 50 gesture
classes from the original dataset. The \texttt{lab\_150} data form the source domain, while the \texttt{home\_276} data form the target domain.
Finally, we use the provided WiMANS~\cite{huang2024wimans} Cross-Env split, whose target domains are an empty room and a meeting room. 

\textbf{Input size.}
In this paper, all methods use CSI amplitude only, without phase information. Each sample is represented as a single-channel amplitude map with 500 temporal samples and 112 CSI dimensions for CSI-Bench, WiAR, and SignFi-50, and 270 CSI dimensions for WiMANS.

\begin{table}[t]
\caption{Dataset statistics and cross-domain information. \# Domains denotes the number of target domains. $N_{\rm src}$ and $N_{\rm tar}$ are the source-training and raw target-split sample counts.}
\centering
\resizebox{\columnwidth}{!}{
\begin{tabular}{lcccc}
\toprule
Dataset & Shift Type & \# Classes & \# Domains & $N_{\rm src}/N_{\rm tar}$ \\
\midrule
CSI-Bench & Cross-Device & 5 & 2 & 21,473 / 9,631 \\
CSI-Bench & Compositional & 5 & 2 & 21,473 / 18,759 \\
WiAR & Cross-User & 16 & 6 & 1,279 / 2,400 \\
SignFi & Cross-Env & 50 & 1 & 1,400 / 500 \\
WiMANS & Cross-Env & 6 & 2 & 1,317 / 3,762 \\
\bottomrule
\end{tabular}}
\label{tab:datasets}
\end{table}

\begin{table*}[ht]
\centering
\caption{Source and one-shot target performance (domain-average Macro-F1). The best results are highlighted in bold.}
\resizebox{\textwidth}{!}{
\begin{tabular}{llccccc|cccc|c}
\toprule
\multirow{2}{*}{\textbf{Dataset} $\downarrow$}
& \multirow{2}{*}{\textbf{Split} $\downarrow$}
& \multicolumn{5}{c}{No Labeled Target Support}
& \multicolumn{5}{c}{One-Shot Target Adaptation} \\
\cmidrule(lr){3-7}\cmidrule(lr){8-12}
&
& ResNet & SimCLR & DANN & CORAL & EI
& PTN & FT & TOSS & FewSense & \textsc{CSI-Agent} \\
\midrule
\multirow{3}{*}{CSI-Bench}
& Source & 0.9810 & 0.8539 & 0.9802 & \textbf{0.9845} & 0.8596 & 0.9503 & 0.9608 & 0.9069 & 0.9618 & 0.9196 \\
& Device & 0.2535 & 0.3403 & 0.2416 & 0.2712 & 0.2535 & 0.2879 & 0.2702 & 0.2275 & 0.2824 & \textbf{0.3464} \\
& Comp.  & 0.2577 & 0.3242 & 0.2725 & 0.2560 & 0.2332 & 0.2383 & 0.2184 & 0.2462 & 0.2238 & \textbf{0.3431} \\
\midrule

\multirow{2}{*}{WiAR}
& Source & 0.8037 & 0.7687 & 0.7795 & 0.8040 & 0.7433 & 0.8567 & 0.8417 & 0.7594 & 0.7030 & \textbf{0.8595} \\
& User   & 0.0418 & 0.0334 & 0.0460 & 0.0484 & 0.0443 & 0.1796 & 0.2702 & 0.0882 & 0.2885 & \textbf{0.4006} \\
\midrule

\multirow{2}{*}{SignFi}
& Source & 0.5984 & 0.7963 & 0.6247 & 0.5863 & 0.6284 & 0.8105 & 0.7994 & 0.6824 & 0.6208 & \textbf{0.8780} \\
& Env.   & 0.0131 & 0.0079 & 0.0033 & 0.0115 & 0.0066 & 0.2926 & 0.2642 & 0.0036 & 0.2873 & \textbf{0.7977} \\
\midrule

\multirow{2}{*}{WiMANS}
& Source & 0.9029 & 0.8195 & 0.9137 & \textbf{0.9195} & 0.9086 & 0.8503 & 0.8616 & 0.8796 & 0.8736 & 0.8948 \\
& Env.   & 0.2687 & 0.2327 & 0.2873 & 0.2821 & 0.3316 & 0.4118 & 0.3896 & 0.4725 & 0.5345 & \textbf{0.5571} \\

\midrule
\multicolumn{2}{l}{\textbf{Target Average}}
& 0.1670 & 0.1877 & 0.1701 & 0.1738 & 0.1738
& 0.2820 & 0.2825 & 0.2076 & 0.3233 & \textbf{0.4890} \\
\bottomrule
\end{tabular}}
\label{tab:main_1shot}
\end{table*}

\textbf{Baselines.}
We compare three categories of cross-domain baselines.
\emph{Source-side representation learning} includes supervised ResNet-18~\cite{he2016deep}, SimCLR pretraining with a linear probe~\cite{chen2020simple}, and EI~\cite{jiang2018towards}.
These methods improve transferability through supervised, self-supervised, or domain-invariant representation learning without using labeled target data.
\emph{Unlabeled target-domain adaptation} includes DANN~\cite{ganin2016domain} and CORAL~\cite{sun2017correlation}, which align source and target feature distributions using unlabeled target observations.
\emph{Few-shot target-domain adaptation} methods use the same $k$-shot support set as CSI-Agent. PTN~\cite{snell2017prototypical} performs classic prototype-based classification, while Fine-Tune (FT) updates the source model using the labeled target samples. FewSense~\cite{yin2022fewsense} adapts the learned representation for similarity-based recognition.
TOSS~\cite{zhou2022target} further combines limited labeled target samples with pseudo-labeled target observations for semi-supervised adaptation.

\subsection{Cross-Domain Performance}

Table~\ref{tab:main_1shot} reports the source-domain and one-shot target-domain performance. Although most methods achieve strong performance on the source domain, their accuracy often drops substantially after deployment. For example, CORAL achieves a Macro-F1 of 0.9845 on the CSI-Bench source domain but only 0.2712 under the Cross-Device split. A similar degradation is observed on WiAR, SignFi, and WiMANS, confirming that strong source-domain recognition does not necessarily translate into robust cross-domain sensing.

The relative performance of existing methods also varies considerably across target domains. SimCLR is the strongest baseline on both CSI-Bench splits, whereas FewSense performs best on WiAR and WiMANS, and PTN achieves the highest baseline result on SignFi. Some methods exhibit even more pronounced target dependence. For example, TOSS reaches 0.4725 on WiMANS but only 0.0036 on SignFi. These results are consistent with our preliminary analysis that no fixed adaptation strategy is uniformly effective across different deployment shifts.

CSI-Agent achieves the best target-domain performance on all five evaluation splits.
On CSI-Bench, it consistently outperforms the strong SimCLR baseline under both device and compositional shifts. The improvements are more substantial under the challenging cross-user and cross-environment settings, where CSI-Agent surpasses the strongest baseline by 0.1121 on WiAR, 0.5051 on SignFi, and 0.0226 on WiMANS. Averaged across the five target splits, CSI-Agent achieves a Macro-F1 of 0.4890, improving upon the strongest baseline method, FewSense, by 0.1657. These consistent results demonstrate the effectiveness of evidence-driven deployment-time adaptation across diverse domain shifts.


\begin{figure}[t]
    \centering
    \includegraphics[width=1\columnwidth]{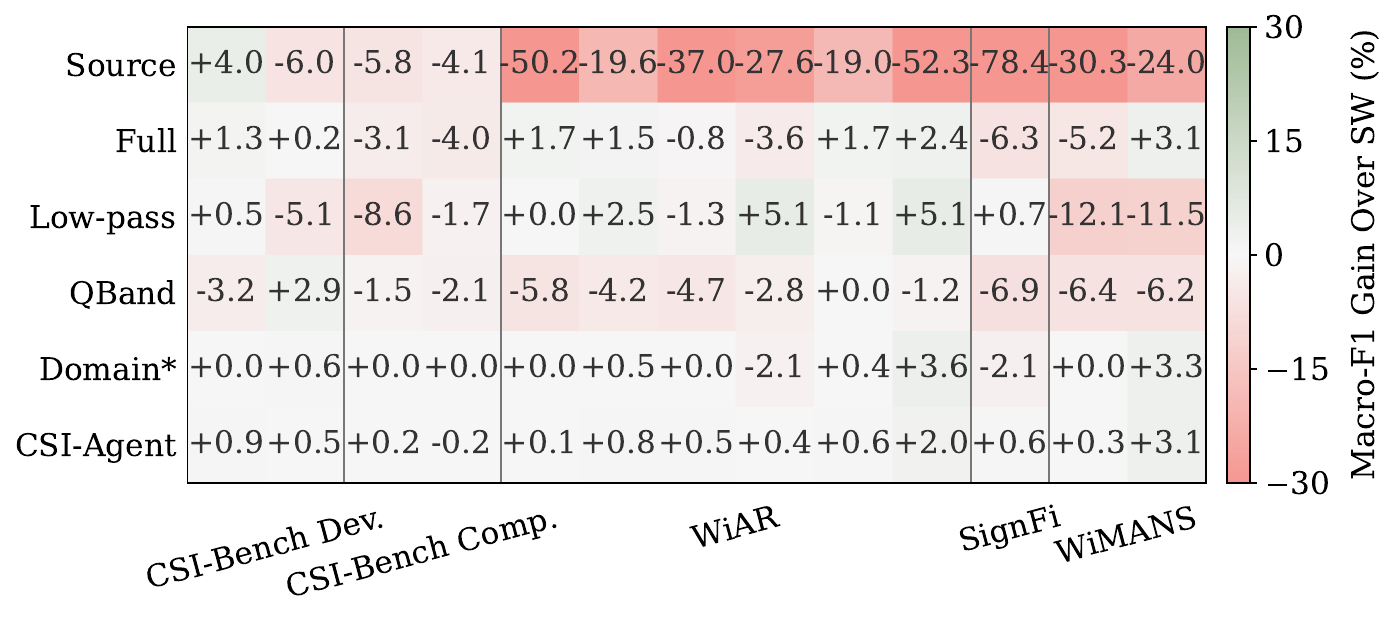}
    \caption{Performance impact of fixed sensing actions and agentic action selection across target domains under 1-shot adaptation. Domain* denotes a domain-level variant of CSI-Agent that selects a single action for all classes in each target domain. Each cell reports the Macro-F1 change in percentage points relative to the support-weighted action, which is omitted from the figure. Positive values indicate improvement, whereas negative values indicate performance degradation.}
    \label{fig:action_utility}
\end{figure}

\subsection{Selective Action Analysis}

\textbf{Target-dependent effects of sensing actions.}
To examine whether the target dependence observed in Section~\ref{sec:pre} extends to the actual action space of CSI-Agent, Fig.~\ref{fig:action_utility} compares each candidate action against the support-weighted default, which provides a strong target-adaptive reference.
The performance impact varies substantially across target domains. For example, the source action improves one CSI-Bench device target by 4\% but causes severe degradation on SignFi, while the Low-pass action improves some WiAR targets but reduces performance on both WiMANS targets. At the same time, Full and Low-pass provide positive gains on several domains and are generally more stable than reverting to the source classifier, indicating that the toolbox contains useful recovery options. 
However, no candidate action is consistently beneficial, and retaining the support-weighted default remains preferable in several domains. These results show that target-dependent effects are not specific to a single few-shot adapter and motivate deployment-time selection.

\textbf{Effectiveness of class-level action selection.}
The bottom two rows of Fig.~\ref{fig:action_utility} compare domain-level and class-level planning under the same support-weighted default. Domain* selects one action for all classes in a target domain, whereas CSI-Agent allows different classes to retain the default or invoke different recovery actions. Domain-level planning provides gains on some targets but remains neutral on many others and causes noticeable degradation in two cases. In contrast, CSI-Agent improves 12 of the 13 target domains, with its only degradation limited to 0.2 percentage points.

CSI-Agent does not always match the best fixed action on every individual domain, since it selects actions without query labels and applies conservative class-level corrections rather than optimizing domain-level test performance.
Nevertheless, it is substantially more consistent than any fixed action. Moreover, CSI-Agent produces positive gains on several targets where none of the individual candidate actions improves over the support-weighted default, showing that class-specific action combinations can recover complementary benefits that are unavailable to global action selection.


\begin{figure}[t]
    \centering
    \begin{minipage}[t]{0.483\columnwidth}
        \centering
        \includegraphics[width=\linewidth]{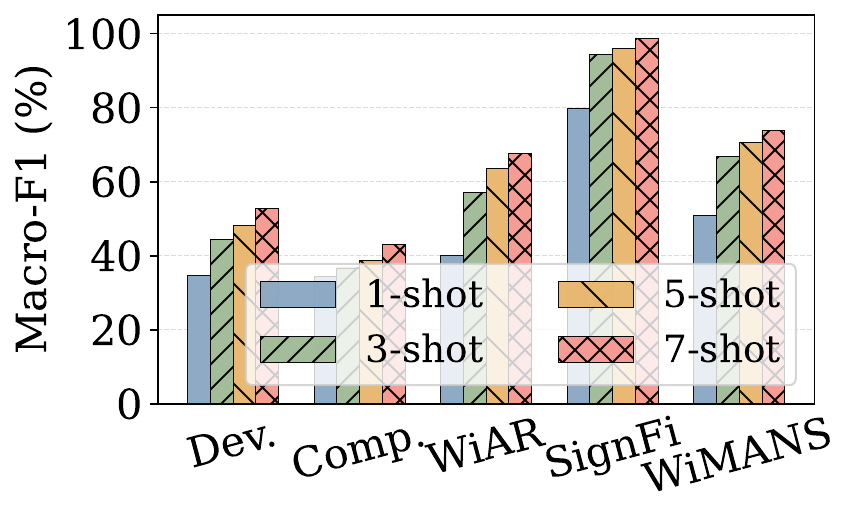}
        \captionof{figure}{Support-size sensitivity of CSI-Agent. Each bar reports Macro-F1 with different $k$ labeled target samples per class. Dev. and Comp. denote the CSI-Bench Cross-Device and Compositional splits.}
        \label{fig:k-shot-sensitivity}
    \end{minipage}
    \hfill
    \begin{minipage}[t]{0.5\columnwidth}
        \centering
        \includegraphics[width=\linewidth]{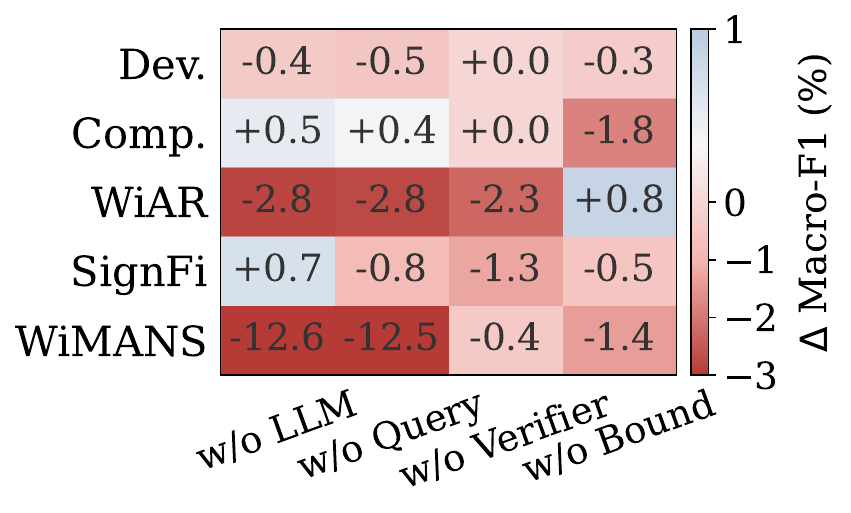}
        \captionof{figure}{Ablation of CSI-Agent under 1-shot Adaptation. Each cell reports the Macro-F1 change relative to the full CSI-Agent. Dev. and Comp. denote the CSI-Bench Cross-Device and Compositional splits.}
        \label{fig:ablation}
    \end{minipage}
\end{figure}

\subsection{Support-Size Sensitivity Analysis}

In practical deployment, $k$ represents the number of labeled target samples collected per sensing class during enrollment. A smaller $k$ reduces the calibration and annotation burden, which is particularly important when target-domain data are difficult to collect.
Fig.~\ref{fig:k-shot-sensitivity} evaluates CSI-Agent with $k\in\{1,3,5,7\}$ labeled samples. Performance improves consistently across all five evaluation cases as additional labeled support becomes available. The largest gains generally occur from 1-shot to 3-shot, particularly on WiAR and WiMANS, showing that even a small amount of additional target supervision can substantially improve adaptation performance. The improvements become more gradual at larger support sizes, with SignFi approaching saturation. 
This trend indicates that CSI-Agent makes effective use of scarce supervision while exhibiting the expected diminishing returns as the support set becomes more representative of the target domain. Importantly, no evaluation split shows unstable or non-monotonic behavior as $k$ increases, suggesting that the sensing actions and diagnostic evidence benefit consistently from additional support.
Overall, CSI-Agent remains effective in the low-cost 1-shot setting while continuing to benefit from additional labeled target samples with diminishing marginal gains.

\subsection{Ablation Study}

Fig.~\ref{fig:ablation} evaluates the contribution of the planning, diagnostic, and safety components compared to the full CSI-Agent under the 1-shot setting. The \emph{w/o LLM} variant replaces the LLM planner with a deterministic support-only rule while retaining the verifier and bounded executor. Its average performance decreases to 45.96\% ($\Delta=-2.93$), with substantial degradation on WiAR and WiMANS. This result shows that selecting actions solely from sparse support-side statistics is insufficient under heterogeneous target shifts.
The \emph{w/o Query} variant retains the LLM planner but removes the diagnostic statistics derived from unlabeled query samples. Its average Macro-F1 drops to 45.65\% ($\Delta=-3.25$), including a pronounced reduction on WiMANS ($\Delta=-12.55$). Together, these results show that reliable action planning requires both query-aware sensing evidence and joint reasoning over class-level diagnostic signals, rather than a fixed support-driven rule.

The verifier and bounded executor provide smaller but important safety benefits. Removing the verifier has little effect on the two CSI-Bench splits but degrades WiAR ($\Delta=-2.3$) and SignFi ($\Delta=-1.3$), indicating that evidence-based proposals can still be unsafe under particular shifts. Removing bounded execution causes clear degradation on CSI-Bench Compositional ($\Delta=-1.8$) and WiMANS ($\Delta=-1.4$). These results show that verification and bounded residual execution complement the planner by limiting the impact of unreliable interventions when they occur.

\subsection{LLM Planner Robustness}

\begin{table}[t]
    \centering
    \caption{Performance of CSI-Agent with different LLM planners under the same adaptation protocol (Macro-F1, \%). Dev. and Comp. denote the CSI-Bench Cross-Device and Compositional splits.}
    \label{tab:llm_type}
    \resizebox{\columnwidth}{!}{
    \begin{tabular}{lcccccc}
        \toprule
        Planner & $k$ & Dev. & Comp. & WiAR & SignFi & WiMANS \\
        \midrule
        Qwen3.5-9B & 1 & 34.64 & 34.31 & 40.06 & 79.77 & 55.71 \\
        Llama-3.1-8B & 1 & 34.70 & 34.16 & 39.89 & 79.77 & 55.65 \\
        GPT-5.4-mini & 1 & 34.40 & 34.44 & 40.03 & 79.41 & 54.55 \\
        Claude Haiku 4.5 & 1 & 33.96 & 34.44 & 40.47 & 79.77 & 55.50 \\
        \midrule
        Qwen3.5-9B & 5 & 48.10 & 38.75 & 63.67 & 95.90 & 67.44 \\
        Llama-3.1-8B & 5 & 48.15 & 38.75 & 63.46 & 95.45 & 67.44 \\
        GPT-5.4-mini & 5 & 48.15 & 38.75 & 63.71 & 95.05 & 67.59 \\
        Claude Haiku 4.5 & 5 & 48.10 & 38.60 & 63.78 & 95.45 & 67.63 \\
        \bottomrule
    \end{tabular}}
\end{table}

Having established that a simple support-driven rule is insufficient, we next examine whether CSI-Agent depends on a particular LLM backbone. We instantiate the planner with Qwen3.5-9B~\cite{qwen35}, Llama-3.1-8B~\cite{grattafiori2024llama}, GPT-5.4-mini~\cite{openai2026gpt54mini}, and Claude Haiku 4.5~\cite{anthropic2025haiku45}, covering both open-weight and proprietary models under the same prompt, evidence, and execution protocol. 
Table~\ref{tab:llm_type} reports their performance under the 1-shot and 5-shot settings. The average Macro-F1 ranges only from 48.57\% to 48.90\% at 1-shot and from 62.65\% to 62.77\% at 5-shot, with no single planner consistently dominating across all target splits. 
These results show that CSI-Agent is robust across the tested LLM planners. The planner makes class-level decisions over structured diagnostic evidence and a small, constrained action space, rather than performing sample-level CSI recognition. Consequently, once an LLM is capable of following the planning protocol and interpreting the diagnostic signals, greater general model capacity does not necessarily translate into better sensing performance. This design allows different capable LLMs to make comparably effective deployment decisions without tying CSI-Agent to a specific backbone.

\subsection{Overhead Analysis}

\begin{table}[t]
    \caption{Deployment overhead per target domain under 1-shot adaptation. Dev. and Comp. denote the CSI-Bench Cross-Device and Compositional splits.}
    \centering
    \scriptsize
    \setlength{\tabcolsep}{2.5pt}
    \renewcommand{\arraystretch}{1.10}
    \resizebox{\columnwidth}{!}{
    \begin{tabular}{lrrrrr}
        \toprule
        \textbf{Configuration $\downarrow$} & Dev. & Comp. & WiAR & SignFi & WiMANS \\
        \midrule
        \multicolumn{6}{l}{\emph{Deployment latency (s/domain)}} \\
        Source Inference
            & 1.28 & 2.40 & 0.14 & 0.37 & 0.99 \\
        Toolbox + Evidence
            & 8.84 & 16.91 & 0.83 & 1.54 & 7.84 \\
        \textsc{CSI-Agent}
            & 20.59 & 28.60 & 33.84 & 91.02 & 21.36 \\
        Sample-Level LLM
            & 7,573.69 & 15,642.78 & 612.37 & 802.78 & 2,752.70 \\
        \addlinespace[1pt]
        \midrule
        \multicolumn{6}{l}{\emph{Input prompt tokens ($10^3$/domain)}} \\
        \textsc{CSI-Agent}
            & 4.81 & 4.85 & 15.19 & 47.67 & 5.90 \\
        Sample-Level LLM
            & 1,643.99 & 3,210.18 & 132.45 & 156.38 & 643.24 \\
        \bottomrule
    \end{tabular}
    \label{tab:overhead}}
\end{table}

CSI-Agent uses the LLM only for class-level deployment planning rather than invoking it separately for individual query samples. Table~\ref{tab:overhead} quantifies the resulting latency and input-token cost. Source inference first processes the target split, while the toolbox constructs the candidate scorers and sensing-grounded diagnostic evidence. The complete CSI-Agent pipeline additionally includes LLM planning, verification, and bounded execution.

The deterministic toolbox and evidence construction introduce moderate overhead beyond source inference. Including class-level LLM planning, the complete pipeline requires 20.59--91.02~s per target domain. The highest cost occurs on SignFi, which contains 50 sensing classes and consequently produces the largest class-level prompt at 47.67K input tokens. This reflects the design of CSI-Agent, whose LLM input scales primarily with the number of classes and candidate actions rather than with the number of individual CSI samples.

We further estimate the cost of a counterfactual sample-level LLM design by constructing compact diagnostic cards for individual queries. Even with these compressed cards instead of raw CSI, sequential sample-level reasoning would require 612.37--15,642.78~s and 132.45K--3.21M input tokens per target domain. Compared with this estimate, CSI-Agent reduces latency by $8.8\times$--$547.0\times$ and input-token cost by $3.3\times$--$661.9\times$.
These results demonstrate that sensing-grounded aggregation makes LLM-assisted adaptation substantially more scalable than sample-level reasoning.

\begin{table}[t]
    \centering
    \caption{Representative evidence-grounded decision traces under 1-shot adaptation. Evidence compares the support-weighted default with the proposed action, while $R_c$ reports the post-hoc recall change of the final bounded output.}
    \label{tab:decision_examples}
    \scriptsize
    \setlength{\tabcolsep}{1.5pt}
    \renewcommand{\arraystretch}{1.1}
    \begin{tabular}{
        L{0.14\columnwidth}
        L{0.25\columnwidth}
        L{0.29\columnwidth}
        L{0.25\columnwidth}}
        \toprule
        Case
        & Decision-Time Evidence
        & Condensed Rationale
        & \makecell[tl]{Decision and\\Outcome} \\
        \midrule
        \makecell[tl]{CSI-Bench\\E05\\ C0}
        &
        $p$: $.136\rightarrow.474$;
        Rank: $2\rightarrow1$;
        
        FA: $.25\rightarrow.25$
        &
        QBand improves support probability and rank without increasing false activation.
        &
        Accept bounded QBand correction.

        $R_c$: $0\rightarrow1$

        $(\Delta R_c=+100~\mathrm{pp})$
        \\
        \midrule
        \makecell[tl]{WiAR\\user-9\\ C7}
        &
        $q$: $.047\rightarrow.056$;
        
        $p$: $.314\rightarrow.367$;
        
        Rank: $2\rightarrow1$;
        
        FA: $.400\rightarrow.333$
        &
        Low-pass moves query activation toward the uniform reference, improves support fit, and reduces false activation.
        &
        Accept bounded Low-pass correction.

        $R_c$: $0\rightarrow1$

        $(\Delta R_c=+100~\mathrm{pp})$
        \\
        \midrule
        \makecell[tl]{SignFi\\home\_276 \\ C0}
        &
        $q$: $.020\rightarrow.020$;
        
        $p$: $.062\rightarrow.089$;
        
        Rank: $4\rightarrow1$;
        
        FA: $.102\rightarrow.020$
        &
        Low-pass preserves query activation near the uniform reference, improves support fit, and reduces false activation.
        &
        Accept bounded Low-pass correction.
        $R_c$: $0\rightarrow1$
        $(\Delta R_c=+100~\mathrm{pp})$
        \\
        \bottomrule
    \end{tabular}
    \parbox{\columnwidth}{\scriptsize
    \emph{Note.} $q$, $p$, Rank, and FA denote query class rate, support probability, support rank, and false activation, respectively.}
\end{table}

\subsection{Illustrative Decision Traces}

Table~\ref{tab:decision_examples} links decision-time evidence with the post-hoc outcome for the same sensing class. The three selected cases illustrate successful non-default interventions under device, user, and environment shifts. QBand recovers a class under the CSI-Bench Cross-Device split, while Low-pass recovers classes under the WiAR cross-user and SignFi cross-environment settings.

The final outcomes are computed using hidden query labels only after the action plan and execution results have been fixed. These labels are unavailable to the diagnostic card, LLM planner, verifier, and bounded executor. The examples therefore illustrate that evidence-grounded decisions can translate into actual class-level recovery rather than providing plausible rationales alone. The reported rationales are condensed from the LLM outputs and serve as auditable decision references rather than faithful causal explanations.

\section{Conclusion}\label{sec:conclusion}

This paper presented CSI-Agent, an evidence-seeking LLM agent for deployment-time adaptation in cross-domain Wi-Fi CSI sensing. Instead of applying one fixed adaptation strategy to an entire target domain, CSI-Agent uses compact, sensing-grounded evidence to diagnose class-specific transfer behavior and determine when specialized recovery actions are justified.
Comprehensive experiments show that CSI-Agent consistently improves target-domain performance across various domain shifts. The results further demonstrate the benefits of class-level action selection, query-aware diagnostic evidence, and constrained planning, while showing robustness across different LLM backbones and avoiding costly sample-level LLM inference. 
These findings suggest that LLM agents can support practical cross-domain sensing when their decisions are grounded in structured sensing evidence and constrained by reliable execution mechanisms.

\bibliographystyle{IEEEtran}
\bibliography{ref}

\end{document}